# Multi-institutional Validation of Two-Streamed Deep Learning Method for Automated Delineation of Esophageal Gross Tumor Volume Using planning-CT and FDG-PET/CT

Xianghua Ye[1†], Dazhou Guo[2†], Chen-kan Tseng[3], Jia Ge[1], Tsung-Min Hung[3], Ping-Ching Pai[3], Yanping Ren[4], Lu Zheng[5], Xinli Zhu[1], Ling Peng[6], Ying Chen[1], Xiaohua Chen[7], Chen-Yu Chou[3], Danni Chen[1], Jiaze Yu[8], Yuzhen Chen[11], Feiran Jiao[9], Yi Xin[10], Lingyun Huang[10], Guotong Xie[10], Jing Xiao[10], Le Lu[2], Senxiang Yan[1]*, Dakai Jin[2]*, Tsung-Ying Ho[11]*

[1]Department of Radiation Oncology, The First Affiliated Hospital Zhejiang University, Hangzhou, China
[2]PAII Inc., Bethesda, MD, USA
[3]Department of Radiation Oncology, Chang Gung Memorial Hospital, Linkou, Taiwan, ROC
[4]Department of Radiation Oncology, Huadong Hospital Affiliated to Fudan University, Shanghai, China
[5]Department of Radiation Oncology, Lihuili Hospital, Ningbo Medical Center, Ningbo, Zhejiang, China
[6]Department of Respiratory Disease, Zhejiang Provincial People's Hospital, Hangzhou, Zhejiang, China
[7]Department of Radiation Oncology, The First Hospital of Lanzhou University, Lanzhou, Gansu, China
[8]Department of Radiation Oncology, Haining People's Hospital, Jiaxing, Zhejiang, China
[9]Statistician, Silver Spring, MD, USA
[10]Ping An Technology, Shenzhen, China
[11]Department of Nuclear Medicine, Chang Gung Memorial Hospital, Linkou, Taiwan, ROC

†Xianghua Ye and Dazhou Guo are co-first authors to this work.
*T-Y. Ho (primary), D. Jin and S. Yan are the joint senior authors to this work.

Emails:
Tsung-Ying Ho: tyho@cgmh.org.tw (Primary contact)
Dakai Jin: dakai.jin@gmail.com (co-corresponding contact)
Senxiang Yan: yansenxiang@zju.edu.cn (co-corresponding contact)

**Title of the manuscript:**

Multi-institutional Validation of Two-Streamed Deep Learning Method for Automated Delineation of Esophageal Gross Tumor Volume using planning-CT and FDG-PET/CT

**Article Type:** Original Research

**Summary statement:**

Deep learning predicted esophageal GTV contours were in close agreement with the ground truth (accuracy comparable to inter-observer agreements), which could increase the efficiency and consistency of radiation oncologists' daily work.

**Key points:**

- A two-streamed deep model achieved high segmentation accuracy in internal testing (Dice score: 0.81 using CT, 0.83 using CT+PET), and generalize well to external evaluation (Dice score: 0.80 using CT).
- Expert's assessment showed that for 88% patients, predicted contours only need minor or no revision.
- Assisted by deep learning, four radiation oncologists' contouring accuracy were substantially improved (averaged Dice score: 0.82 vs 0.84, $p<0.001$); and inter-observer variation and contouring time were reduced by 37.6% and 48.0%, respectively.

**Abbreviations:**

pCT = treatment planning CT, GTV = gross tumor volume, DSC = Dice score, ASD = average surface distance, HD95 = 95% Hausdorff distance, cT2 = clinical T-stage 2, cT3 = clinical T-stage 3, and cT4 = clinical T-stage 4.

**Abstract**

**Background**: The current clinical workflow for esophageal gross tumor volume (GTV) contouring relies on manual delineation of high labor-costs and inter-user variability.

**Purpose**: To validate the clinical applicability of a deep learning multi-modality esophageal GTV contouring model, developed at one institution whereas tested at multiple institutions.

**Methods and Materials**: We collected 606 patients with esophageal cancer retrospectively from four institutions. 252 patients from institution-1 contained both a treatment planning-CT (pCT) and a pair of diagnostic FDG-PET/CT; 354 patients from other three institutions had only pCT scans under different staging protocols or lacking PET scanners. A two-streamed deep learning model for GTV segmentation was developed using pCT and PET/CT scans of a subset (148 patients) from institution-1. This built model had the flexibility of segmenting GTVs via only pCT, or pCT+PET/CT combined when available. For independent evaluation, the rest 104 patients from institution-1 behaved as unseen internal testing, and 354 patients from other three institutions were used for external testing. Degrees of manual revision were further evaluated by human experts to assess the contour-editing effort. Furthermore, the deep model's performance was compared against four radiation oncologists in a multi-user study using 20 randomly chosen external patients. Contouring accuracy and time were recorded for the pre- and post-deep learning assisted delineation process.

**Results**: Our two-streamed deep model achieved high segmentation accuracy in internal testing (mean Dice score (DSC): 0.81 using pCT and 0.83 using pCT+PET) and generalized well to external evaluation (mean DSC: 0.80 using pCT). Experts' assessment showed that the predicted contours of 88% patients need only minor or no revision. In multi-user evaluation, with the assistance of deep model, inter-observer variation and required contouring time were reduced by 37.6% and 48.0%, respectively.

**Conclusions**: Deep learning predicted GTV contours were in close agreement with the ground truth and could be adopted clinically with mostly minor or no changes.

## Introduction

Gross tumor volume (GTV) contouring is an essential task in radiotherapy planning. GTV refers to the demonstrable gross tumor region. Accurate contouring improves patient prognosis and serves as the basis for further clinical target volume delineation (1). For precise GTV delineation, radiation oncologists often need to consider multi-modality imaging of MRI, FDG-PET, contrast-enhanced CT, and radiology reports and other relevant clinical information. This manual process is both labor-intensive and highly variable.

For esophageal cancer, concurrent chemoradiation therapy is the recommended primary treatment, as relatively fewer patients are first diagnosed at asymptomatic early stages eligible for esophagostomy. Compared to other types of cancers, esophageal GTV contouring has its unique challenges: (a) esophagus possesses a long cranial to caudal anatomical range, where tumors may appear at any locations along this tubular organ. Multifocal tumors are also not uncommon (2, 3). Accurately identifying the tumor location needs significant efforts and expertise from radiation oncologists by referring to multiple examinations, such as panendoscopy, contrast esophagography, or FDG-PET/CT. (b) Assessing the longitudinal esophageal tumor extension is difficult on CT, even with additional information from PET. This leads to considerable GTV contouring variations at the cranial-caudal border (4, 5). (c) Treatment planning-CT (pCT) exhibits poor contrast between the esophageal tumor and surrounding tissues. This limitation is addressed by frequently manual referring to adjacent slices to delineate GTV's radial borders, further increasing the manual burden and time. Therefore, current manual esophageal GTV contouring is labor-intensive and requires extensive experiences of radiation oncologists, otherwise leading to inconsistent delineation. Accurate and automated GTV contouring is of evidently great benefits.

Deep learning methods have been demonstrated as potentially clinical relevant and useful tools in many medical image analysis tasks (6-8). The deep learning-based target volume and organ at risk contouring were also increasingly studied recently (9-14). Nevertheless, for esophageal GTV, the clinical applicability of deep learning-based auto-contouring is still unclear under a multi-institutional evaluation setup.

In this study, we developed and validated a two-streamed 3D deep learning esophageal GTV segmentation model, which had the flexibility to segment the GTV using only pCT, or pCT and PET/CT combined when available. The deep model was developed using 148 patients with pCT and PET/CT imaging from institution-1, and independently validated using 104 unseen patients from institution-1 and 354 patients from three external institutions. Furthermore, using 20 randomly selected patients from external institutions, the deep model performance was compared under a multi-user setting with four board-certified radiation oncologists experienced in esophageal cancer treatment.

## Materials and Methods

*Datasets*

Total 606 patients with esophageal cancer from four institutions were collected in this retrospective study under each institutional review board approval; requirements to obtain informed consent were waived. All patients had undergone concurrent chemoradiation therapy before surgery between 2015 and 2020. The exclusion criteria are shown in Fig. 1. All patients had available pCT scans and the corresponding manual GTV contours used for clinical treatment. According to the availability of PET scanner and the staging protocol of different institutions, patients from institution-1 (252 patients total) received additional diagnostic FDG-PET/CT scan, whereas 354 patients from other institutions collected only pCT. Imaging details are described in Appendix.  A subset of 148 patients from institution-1 was used as the training/validation cohort, while the rest 104 patients from institution-1 and 354 patients from other three institutions were treated as unseen internal and external testing cohorts, respectively (Fig. 1). 148 (institution-1) of the 606 patients were previously reported (15). This prior work dealt with segmentation method developments, whereas in this manuscript we constructed the deep model using different implementation (Appendix) and evaluated the performance on 458 unseen multi-institutional patients (104 from institution-1, 354 from other three institutions).

*Model development*

We implemented a two-streamed 3D esophageal GTV segmentation method based on the process described in (15), which consisted of a pCT stream to segment GTVs using only pCT input (denoted pCT model), and a pCT+PET stream using an early fusion module followed by a late fusion module to segment GTVs leveraging the joint information in pCT and PET multi-modalities (denoted pCT+PET model). The overall segmentation flowchart is illustrated in Fig. 2. In the pCT+PET stream, PET was aligned to pCT by first registering the diagnostic CT (accompanying the PET) to pCT and applying the deformation field to map PET to pCT. For segmentation backbone, 3D progressive semantically nested network (15) was adopted. Details of the registration, two-streamed formulation, and network architecture are included in Appendix.

To obtain the final models for testing, we conducted four-fold cross-validation (split at the patient level) on the 148 training-validation patients from institution-1. Thereby, 148 patients were randomly partitioned into four equal-sized subgroups (25% of patients). Of the four subgroups, a single subgroup was retained as the validation data for model selection, while the remaining three subgroups were used for training. The cross-validation process was repeated four times/folds, with each of the four subgroups used once as the validation data. Finally, four deep models were obtained from the four rounds of training. They were ensembled to predict the final GTV contours in all the unseen testing data.

*Quantitative Evaluation of Contour Accuracy*

The contouring accuracy was quantitatively evaluated using three common segmentation metrics (9, 10), i.e., Dice similarity coefficient (DSC), 95% Hausdorff distance (HD95) and average surface distance (ASD). For the internal testing, the performance of pCT or pCT+PET model was separately computed. During external testing, the pCT model performance was reported. We also explored the comparison of these metrics in subgroups with different characteristics: clinical T-stages and different tumor locations (cervical, and upper, middle and lower third of esophagus according to (16)).

Additionally, the performance of our two-streamed models was compared with the previous state-of-the-art method (17), using a 3D denseUnet (18, 19) for pCT-based esophageal GTV segmentation. For the model development of (17), the same four-fold cross-validation protocol were applied to ensure a neutral comparison.

*Human Experts' Assessment of Contour Accuracy*

An assessment experiment by human experts was further conducted to evaluate the contour editing efforts required for deep model predictions to be clinically accepted. Specifically, deep learning predictions of 354 patients from three external multi-institutions were distributed to two experts (both >15 years of experience) to assess the degree of manual revision that was defined as the percentage of GTV slices needed modification for clinical acceptance. Five categories were designated as no revision required, revision required in <10% slices, revision required in 10% to 30% slices, revision required in 30% to 60% slices and unacceptable (revision required in >60% slices or prediction completely missed the tumor). We analyzed the correlations between different quantitative metrics and degrees of manual revision.

Note that esophageal GTV may appear at any esophageal location spanning an extensive longitudinal range, which is different from the more spatially constrained anatomical location such as head & neck or prostate (9, 10). Hence, automated esophageal GTV segmentation may identify wrong tumor locations. These scenarios could lead to large or undefined distance errors. Therefore, for the quantitative evaluation, we additionally report the number of patients identified as unacceptable by clinical experts, and calculated the DSC, HD95 and ASD metrics using the rest patients.

*Multi-user Evaluation*

We further conducted a multi-user study involving four board-certified radiation oncologists (3-6 years experience in treating esophageal cancer) from 4 different institutions. First, pCT of 20 randomly selected patients in the external testing cohort along with their clinical, radiological and panendoscopy reports, and any other useful information were extracted and provided to these four radiation oncologists for manual contouring. Next, after a minimum interval of one-month, deep model predicted GTV contours were distributed to these four radiation oncologists for editing along with previously available information. All radiation oncologists were blinded to the ground truth contours and their first-time contours. Accuracy of our deep model predictions was compared to the multi-user performance in terms of DSC, HD95 and ASD. Similar to (9), inter-observer variations were assessed using multi-user DSC and volume coefficient of variation (the ratio between standard deviation and mean). Time used for the pre- and post-deep learning assisted contouring were recorded.

*Statistical Analysis*

The Wilcoxon matched-pairs signed rank test was used to compare (1) DSC, HD95 and ASD scores between the pCT model and pCT+PET model in the internal testing set; and between the proposed model and 3D DenseUNet method in the external testing set; (2) DSC, HD95, ASD, and time taken of pre- versus post-deep learning assisted contouring in multi-user evaluation. Manning-Whitney U test was used to compare DSC, HD95 and ASD at different clinical T-stages. Multiple linear regression with stepwise model selection was used to compare the metrics at different tumor locations, since a large tumor may locate across multiple esophagus regions. Spearman correlation coefficients were assessed for mean DSC, HD95 and ASD versus degrees of manual revision, respectively. The $\chi^2$ test was used to compare the difference in degrees of manual revision between subgroups. All analyses were performed by using **R** (20). Statistical significance was set at two-tailed *p*<0.05.

## Results

A total of 606 esophageal cancer patients were included. Table 1 summarizes main characteristics of the entire cohort, and the separated training-validation, internal testing and external testing cohorts. Characteristics of the 20 randomly selected patients used in multi-user evaluation are presented in Appendix Table A1.

*Performance in the internal testing set*

Quantitative performance of our deep model in the internal testing set is summarized in Table 2 and Table 3. For the pCT model, we observed the mean and 95% confidence interval of DSC, HD95 and ASD as 0.81 (0.79, 0.83), 11.5 (9.2, 13.7)mm and 2.7 (2.2, 3.3)mm, respectively. In the subgroup analysis (Appendix Fig. A3, A4), the pCT model achieved a significantly higher mean DSC for advanced T-stage patients (cT3, cT4) than those in the early cT2 patients (0.82 and 0.82 versus 0.76, *p*<0.05). The tumor locations exhibited no significant performance differences. With additional PET scans, the pCT+PET model significantly increased the performance to 0.83 (0.81, 0.84), 9.5 (8.0, 10.9)mm and 2.2 (1.9, 2.5)mm with *p*<0.01 in DSC, HD95 and ASD, respectively. Figure 4-(a) shows several qualitative examples for GTV segmentation in the internal testing set.

*Performance in the external testing set*

In the external multi-institutional testing, we observed the mean and 95% CI of DSC, HD95 and ASD as 0.80 (0.78, 0.81), 11.8 (10.1, 13.4)mm and 2.8 (2.4, 3.2)mm, respectively, using the pCT model (Table 4). These values did not show significant differences compared to those during the internal testing. Our pCT-based GTV segmentation model generalized well to patients of three other institutions. In the subgroup analysis, similar trend was observed as internal testing: deep model obtained markedly improved DSC and HD95 in advanced cT3 and cT4 patients versus early cT2 patients (mean DSC 0.81 and 0.82 versus 0.71, $p<0.001$; mean HD95 11.4 and 11.4mm versus 13.8mm, $p\leq0.001$).

When compared with the previous leading 3D DenseUNet (17), its DSC, HD95 and ASD scores were all inferior to our model performance, e.g., mean DSC 0.75 vs 0.80, $p<0.001$ (Appendix Table A2).

*Human experts' Assessment*

Human experts' assessment showed that the majority (311 of 354, 88%) of deep learning predictions in the external testing set were clinically accepted or required only minor editing (no revision n=220; 0-30% revision, n=91). Ten (3%) patients had contouring errors in 30%-60% slices and 33 (9%) patients had unacceptable predictions that required substantial editing efforts. Fig. 4 details the assessment results. The mean DSC, HD95 and ASD were correlated to the degrees of manual revision (DSC: R=-0.58, $p<0.001$; HD95: R=0.60, $p<0.001$; ASD: R=0.60, $p<0.001$). These results indicated the reliability of using DSC, HD95 and ASD as contouring accuracy evaluation criteria, consistent with the contour editing effort necessitated in actual clinical practice. 33 (9%) patients had unacceptable predictions where our model failed to accurately locate the tumor, leading to small dice and large distance errors. Among 33 unaccepted cases, 23 (70%) patients had cT2 tumors. Other cases often exhibited relatively uncommon scanning position or anatomies (see the most-right column in Fig. 3 (b)). In the subgroup analysis (Appendix Table A3), a significantly smaller percentage of patients required major revision (>30% slice revision) in advanced cT3 and cT4 stages as compared to that in early cT2 stage (5% and 8% vs 35%, $p$<0.01). Tumor locations did not significantly impact the degrees of manual revision.

*Multi-user evaluation*

Performance evaluation of our pCT model with four board-certified radiation oncologists is shown in Fig. 5 and Appendix Table A4. Among 20 testing cases, our model performed comparably to these four radiation oncologists in terms of DSC and ASD (mean DSC: 0.82 vs 0.82, 0.83, 0.79, 0.82; mean ASD: 2.0mm vs 1.9, 1.8, 2.6, 2.0mm). For HD95, our model achieved the lowest mean HD95 errors among all results (significantly smaller than R3, mean HD95 7.9mm vs 12.0mm, $p$=0.01).

Next, we examined if the accuracy of manual contouring could be improved with assistance of deep model predictions. It is observed that when editing upon deep model predictions, 2 out of 4 radiation oncologists' performance had been significantly improved in DSC and HD95 (Fig. 5, Appendix Table A5). The inter-user variation was also reduced (Fig. 6). Mean multi-user DSC was improved from 0.82 to 0.84 ($p$<0.001) and mean volume coefficient of variation was reduced by 37.6% (from 0.14 to 0.09, $p$=0.03). Furthermore, the contouring time had been reduced by an average of 48.0% (from 10.2 to 5.3 minutes). Our pCT model takes an average of 20 seconds to predict one patient.

## Discussion

In this multi-institutional study, we developed a two-streamed 3D deep learning model to segment esophageal gross tumor volume (GTV) trained on 148 patients with both treatment planning CT (pCT) and PET/CT scans from institution-1. The performance was extensively evaluated using 104 unseen institution-1 patients and 354 external multi-institutional patients. Our pCT model achieved mean Dice similarity coefficient (DSC) and average surface distance (ASD) of 0.81 and 2.7mm in the internal testing and generalized well to the external testing with mean DSC and ASD of 0.80 and 2.8mm. Adding PET scans, the pCT+PET model further boosted DSC and ASD to 0.83 and 2.2mm for the internal testing. From a multi-user study, the pCT model performed favorably when compared against four board-certified radiation oncologists in metrics of DSC and ASD, while achieving the smallest HD95 errors. By allowing radiation oncologists to edit the deep model predictions, the overall accuracy was improved, and inter-observer variation and contouring time were reduced by 37.6% and 48.0%, respectively.

Accurate GTV delineation improves patient's prognosis (1). Manual contouring of esophageal GTV on pCT highly relies on the expertise and experiences of radiation oncologist leading to substantial inter-user variations (4, 5, 21). In clinical practice, radiation oncologists almost always need to refer to other information such as panendoscopy report to determine the tumor range, which is not trivial requiring the "virtual fusion" of panendoscopy information with pCT image in their minds. In this context, our deep model could benefit the radiation oncologists by improving their contouring accuracy and consistency, and reducing time spent.

Previous works showed potential clinical applicability of deep learning for the GTV contouring in head & neck and prostate cancers (9, 10). However, for esophageal cancer, studies often collected limited single institutional data for both training and testing (15, 17, 22). In this work, with our deep model developed using 148 patients from the internal institution-1, we extensively evaluated the GTV segmentation performance using 104 unseen internal patients and 354 external multi-institutional patients. Robust performance generalizability to the external multi-institutional testing data was observed despite variations of CT scanner types, imaging protocols and patient populations.

Generalizability of deep learning models was often the bottleneck for successful clinical deployment. As shown in (23), direct deployment of well-trained MRI-based prostate and left arterial segmentation models to the unseen data from different centers led to averaged > 10% DSC decrease. Good generalizability of our model may come from: (1) relative standardized imaging protocols for pCT from various institutions despite different pCT scanner vendors; (2) physically well-calibrated HU values in CT; (3) the 148 training patients from institution 1 are relatively sufficient for covering different CT characteristics of esophageal tumors and (4) we have effectively trained our two-streamed deep GTV networks.

The developed two-streamed model has demonstrated the flexibility of segmenting esophageal GTV according to the availability of PET/CT scans. For institutions where PET/CT scans are not included as a standardized staging protocol, our pCT model already achieved high accuracy comparable to the inter-user agreement. When PET/CT scans were available, the pCT+PET model could further improve the performance (mean DSC of pCT vs pCT+PET: 0.81 vs 0.83, $p$=0.01).

This study has a few limitations. First, patients in the external test set do not have PET/CT scans because PET is either not available or not required for esophageal cancer staging in three external institutions. Hence, we have not directly validated the performance of our pCT+PET model in the external data. However, considering that tumor contrast in PET is often prominent and can be assessed as a semi-quantitative standard uptake value (SUV), we believe that it would not significantly impact our pCT+PET model performance when applied to external patients. Second, the pCT model obtained lower performance for patients of cT2 as compared to those of advanced clinical T-stages. This may be because cT2 tumors often exhibited less prominent imaging features in CT. After adding PET, this phenomenon was less evident. Another potential solution might be combining the panendoscopy report information with a deep learning model, which could be optimized by restricting longitudinal ranges. Third, we excluded patients with the primary tumor at gastroesophageal junction, since they were relatively rare (< 2%) in our study population and some were treated by surgery. Further investigation of developing the deep learning model on this sub-population would be of clinical interests.

To conclude, we developed and clinically validated an effective two-streamed 3D deep model that can reliably segment the esophageal GTV using two protocols of pCT alone or pCT+PET/CT. Predicted GTV contours for 88% patients were in close agreement with the ground truth and could be implemented and adopted clinically where only minor or no editing efforts are required.

Table 1. Subject and imaging characteristics.

| Characteristics | Entire cohort Institutions 1-4 (n = 606) | Training-validation Institution 1 (n = 148) | Internal testing Institution 1 (n = 104) | External testing Institutions 2-4 (n = 354) |
|---|---|---|---|---|
| Sex | … | … | … | … |
|   Male | 537 (89%) | 135 (91%) | 98 (94%) | 304 (86%) |
|   Female | 69 (11%) | 13 (9%) | 6 (6%) | 50 (14%) |
| Diagnostic age | 65 [57-72] | 55 [50-61] | 56 [50-62] | 67 [61-75] |
| Clinical T stage | … | … | … | … |
|   cT2 | 116 (19%) | 24 (16%) | 18 (17%) | 74 (21%) |
|   cT3 | 306 (51%) | 71 (48%) | 58 (56%) | 177 (50%) |
|   cT4 | 184 (30%) | 53 (36%) | 28 (27%) | 103 (29%) |
| Tumor location | … | … | … | … |
|   Cervical | 81 (13%) | 11 (7%) | 10 (10%) | 60 (17%) |
|   Upper third | 204 (34%) | 26 (18%) | 35 (34%) | 143 (40%) |
|   Middle third | 325 (54%) | 84 (57%) | 63 (61%) | 178 (50%) |
|   Lower third | 174 (29%) | 69 (47%) | 35 (34%) | 70 (20%) |
| BMI | … | … | … | … |
|   < 18.5 | 121 (20%) | 22 (15%) | 15 (14%) | 84 (24%) |
|   18.5 – 23.9 | 393 (65%) | 94 (63%) | 59 (57%) | 240 (68%) |
|   > 24 | 92 (15%) | 32 (22%) | 30 (29%) | 30 (8%) |
| Imaging available | … | … | … | … |
|   pCT | 606 (100%) | 148 (100%) | 104 (100%) | 354 (100%) |
|   PET/CT | 252 (42%) | 148 (100%) | 104 (100%) | 0 (0%) |

Note: patients may have tumors located across multiple esophagus region, hence, total numbers summed at various tumor locations for the entire and sub-institution cohorts are greater than the corresponding total patient numbers. Age is presented as median and [interquartile range]. cT2 = clinical T-stage 2, cT3 = clinical T-stage 3, and cT4 = clinical T-stage 4. BMI = body mass index. pCT = treatment planning CT.

Table 2. Quantitative results of esophageal GTV segmentation by the pCT model in the unseen internal testing data.

|  | Institution 1 (unseen internal testing) using pCT model | | | |
| --- | --- | --- | --- | --- |
| Evaluation metrics | unacceptable number (percentage) | DSC mean (95% CI) | HD95 (mm) mean (95% CI) | ASD (mm) mean (95% CI) |
| Total patients (n=104) | 8 (8%) | 0.81 (0.79, 0.83) | 11.5 (9.2, 13.7) | 2.7 (2.2, 3.3) |
| Clinical T-stage | | | | |
| cT2 (n=18) | 4 (22%) | 0.76 (0.67, 0.86) | 12.0 (5.5, 18.4) | 3.0 (1.0, 5.1) |
| cT3 (n=58) | 3 (5%) | 0.82 (0.80, 0.84) | 10.7 (7.9, 13.5) | 2.5 (1.9, 3.2) |
| cT4 (n=28) | 1 (4%) | 0.82 (0.79, 0.85) | 12.8 (7.9, 17.7) | 3.0 (2.0, 4.0) |
| Tumor location | | | | |
| Cervical (n=10) | 1 (10%) | 0.82 (0.75, 0.89) | 9.2 (6.5, 12.0) | 2.2 (1.5, 2.8) |
| Upper third (n=35) | 1 (3%) | 0.83 (0.81, 0.85) | 9.6 (7.4, 11.9) | 2.2 (1.8, 2.5) |
| Middle third (n=63) | 5 (8%) | 0.80 (0.78, 0.83) | 12.0 (8.9, 15.0) | 2.9 (2.2, 3.6) |
| Lower third (n=35) | 2 (6%) | 0.81 (0.77, 0.85) | 13.3 (8.6, 18.0) | 3.3 (2.1, 4.5) |

Note: CI = confidence interval, DSC = Dice similarity coefficient, HD95 = 95% Hausdorff distance, ASD = average surface distance, cT2 = clinical T-stage 2, cT3 = clinical T-stage 3, and cT4 = clinical T-stage 4. pCT = treatment planning CT.

Table 3. Quantitative results of esophageal GTV segmentation by the pCT+PET model in the unseen internal testing data.

| Evaluation metrics | Institution 1 (unseen internal testing) using pCT+PET model | | | |
| --- | --- | --- | --- | --- |
| | unacceptable number (percentage) | DSC mean (95% CI) | HD95 (mm) mean (95% CI) | ASD (mm) mean (95% CI) |
| Total patients (n=104) | 4 (4%) | 0.83 (0.81, 0.84) | 9.5 (8.0, 10.9) | 2.2 (1.9, 2.5) |
| Clinical T-stage | | | | |
| cT2 (n=18) | 3 (17%) | 0.77 (0.69, 0.85) | 11.4 (6.3, 16.6) | 2.7 (1.3, 4.2) |
| cT3 (n=58) | 0 (0%) | 0.84 (0.82, 0.85) | 9.0 (7.0, 11.0) | 2.0 (1.7, 2.4) |
| cT4 (n=28) | 1 (4%) | 0.84 (0.82, 0.86) | 9.3 (7.3, 11.4) | 2.3 (1.9, 2.6) |
| Tumor location | | | | |
| Cervical (n=10) | 1 (10%) | 0.83 (0.78, 0.89) | 9.4 (6.2, 12.7) | 2.0 (1.5, 2.5) |
| Upper third (n=35) | 0 (0%) | 0.84 (0.82, 0.86) | 8.1 (6.2, 10.0) | 1.9 (1.6, 2.2) |
| Middle third (n=63) | 3 (5%) | 0.83 (0.81, 0.84) | 9.5 (8.0, 11.1) | 2.2 (1.9, 2.5) |
| Lower third (n=35) | 0 (0%) | 0.83 (0.79, 0.86) | 10.8 (7.5, 14.0) | 2.6 (1.9, 3.3) |

Note: CI = confidence interval, DSC = Dice similarity coefficient, HD95 = 95% Hausdorff distance, ASD = average surface distance, cT2 = clinical T-stage 2, cT3 = clinical T-stage 3, and cT4 = clinical T-stage 4. pCT = treatment planning CT.

Table 4. Quantitative results of esophageal GTV segmentation by the pCT model in the multi-institutional external testing data.

| Evaluation metrics | Institution 2-4 (external multi-institutional testing) using pCT model | | | |
| --- | --- | --- | --- | --- |
| | unacceptable number (percentage) | DSC mean (95% CI) | HD95 (mm) mean (95% CI) | ASD (mm) mean (95% CI) |
| Total patients (n=354) | 33 (9%) | 0.80 (0.78, 0.81) | 11.8 (10.1, 13.4) | 2.8 (2.4, 3.2) |
| Clinical T stage | | | | |
| cT2 (n=74) | 23 (31%) | 0.71 (0.66, 0.76) | 13.8 (10.0, 17.5) | 3.6 (2.5, 4.8) |
| cT3 (n=177) | 5 (3%) | 0.81 (0.80, 0.82) | 11.4 (8.8, 13.9) | 2.6 (2.1, 3.2) |
| cT4 (n=103) | 5 (5%) | 0.82 (0.80, 0.83) | 11.4 (9.3, 13.6) | 2.7 (2.1, 3.3) |
| Tumor location | | | | |
| Cervical (n=60) | 4 (6%) | 0.80 (0.78, 0.82) | 11.7 (8.6, 14.8) | 2.5 (1.7, 3.3) |
| Upper third (n=143) | 11 (8%) | 0.79 (0.77, 0.81) | 12.6 (10.4, 14.9) | 3.0 (2.4, 3.7) |
| Middle third (n=178) | 14 (8%) | 0.80 (0.78, 0.81) | 11.5 (9.3, 13.5) | 2.9 (2.4, 3.5) |
| Lower third (n=70) | 5 (7%) | 0.80 (0.78, 0.82) | 15.4 (9.3, 21.5) | 3.3 (2.1, 4.5) |

Note: CI = confidence interval, DSC = Dice similarity coefficient, HD95 = 95% Hausdorff distance, ASD = average surface distance, cT2 = clinical T-stage 2, cT3 = clinical T-stage 3, and cT4 = clinical T-stage 4. pCT = treatment planning CT.

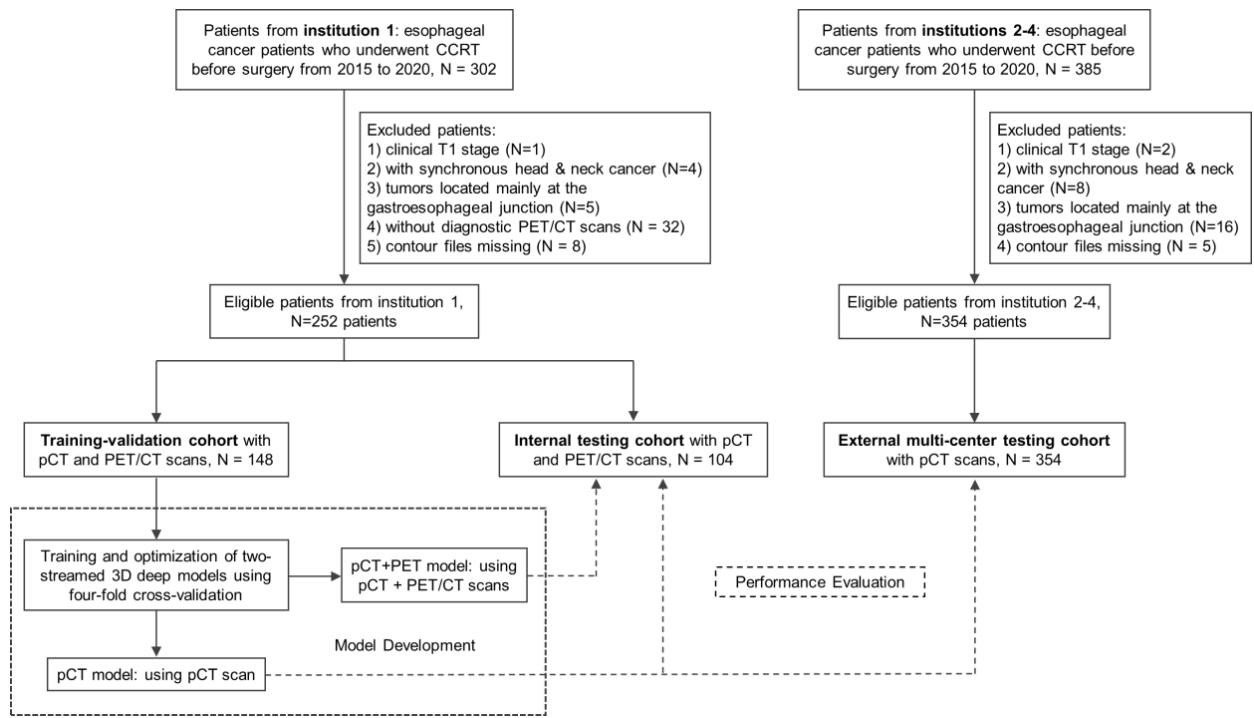

Fig. 1. Study flow diagram of esophageal gross tumor volume (GTV) segmentation in a multi-institutional setup. CCRT = concurrent chemoradiation therapy, pCT = treatment planning CT.

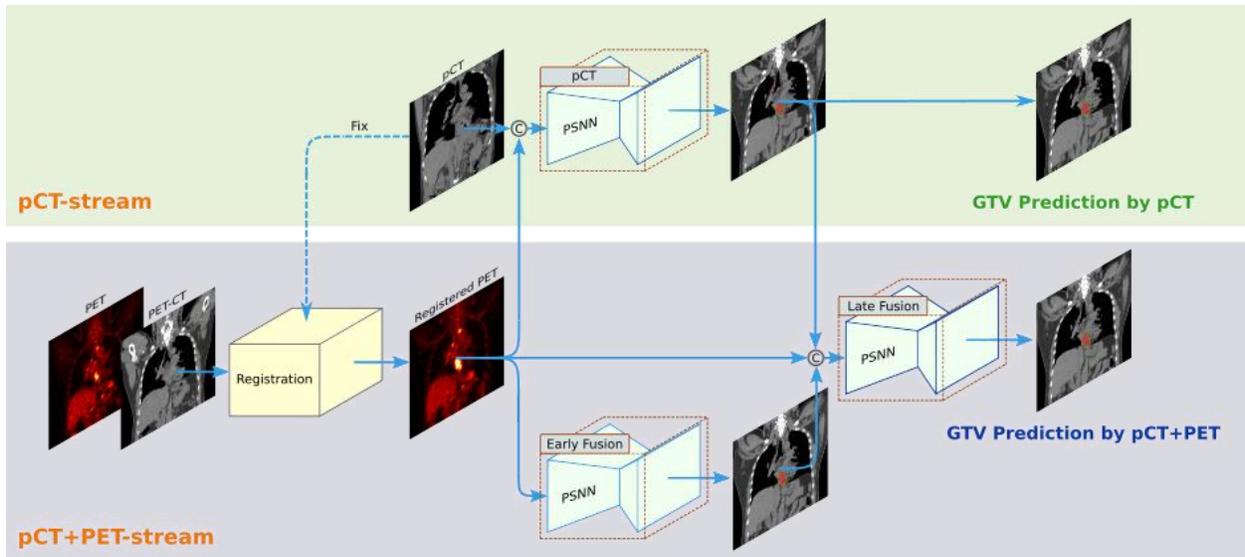

Fig. 2. The two-streamed 3D deep learning model for esophageal gross tumor volume (GTV) segmentation using treatment planning CT (pCT) and FDG-PET/CT scans. pCT stream takes the pCT as input and produces the GTV segmentation prediction. The pCT+PET stream takes both pCT and PET/CT scans as input. It first aligns the PET to pCT by registering diagnostic CT (accompanying PET scan) to pCT and applying the deformation field to further map PET to pCT. Then, it uses an early fusion module followed by a late fusion module to segment the esophageal GTV using the complementary information in pCT and PET. This workflow can accommodate to the availability of PET scans in different institutions. Although 3D inputs are used, we depict 2D images for clarity.

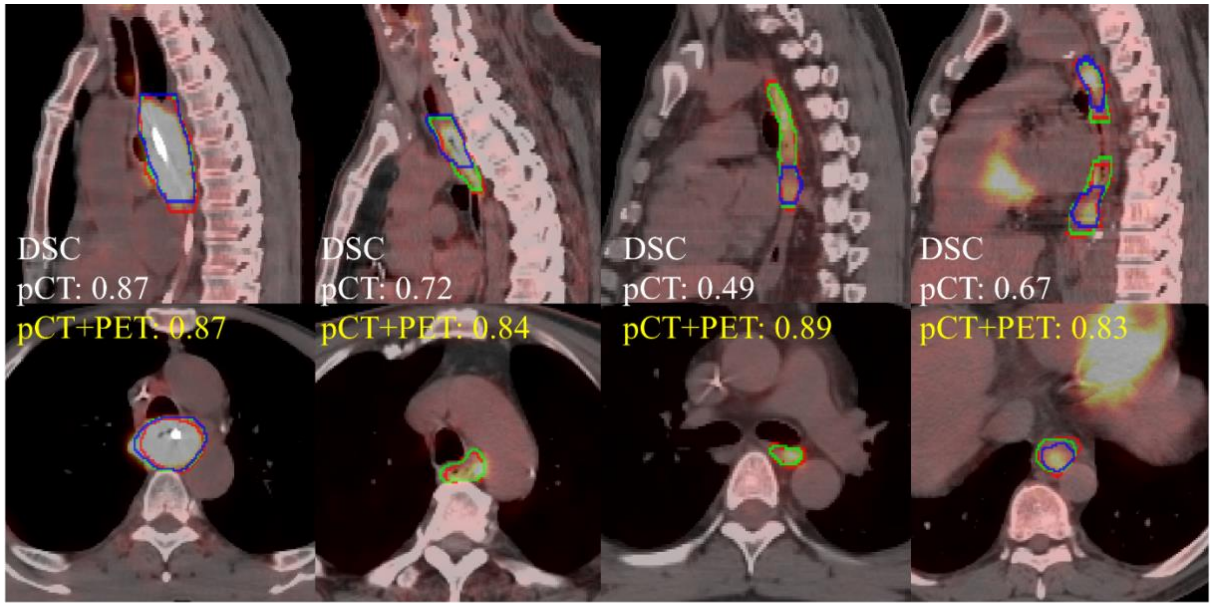

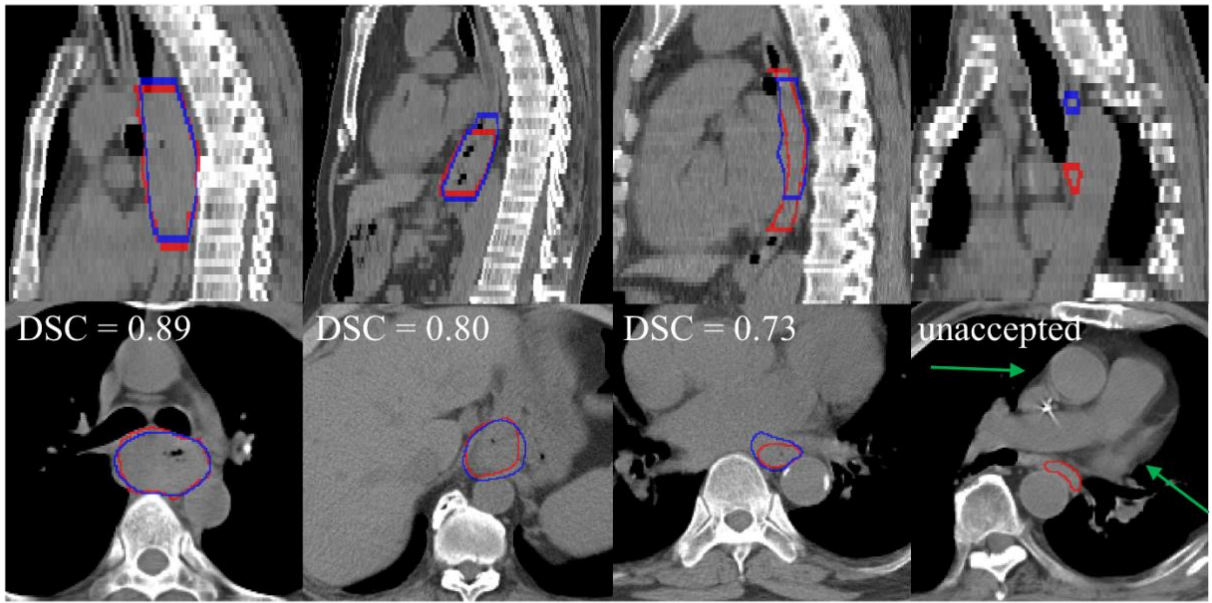

Fig. 3. (a): Performance comparison of pCT model and pCT+PET in the internal testing set (left to right: cT4, cT3, cT3, multifocal cT2). Red, green and blue color show the contours of ground truth reference, pCT+PET model prediction, and pCT model prediction. (b) Performance examples of pCT model in the external testing set according to the degree of manual revision (left to right): no revision, >0-30%, >30-60%, and unacceptable. Red and blue color show the contours of ground truth reference and pCT model prediction, respectively. Green arrow points to the uncommon anatomy for the unacceptable case in the most-right column. pCT = treatment planning CT, DSC = Dice similarity coefficient.

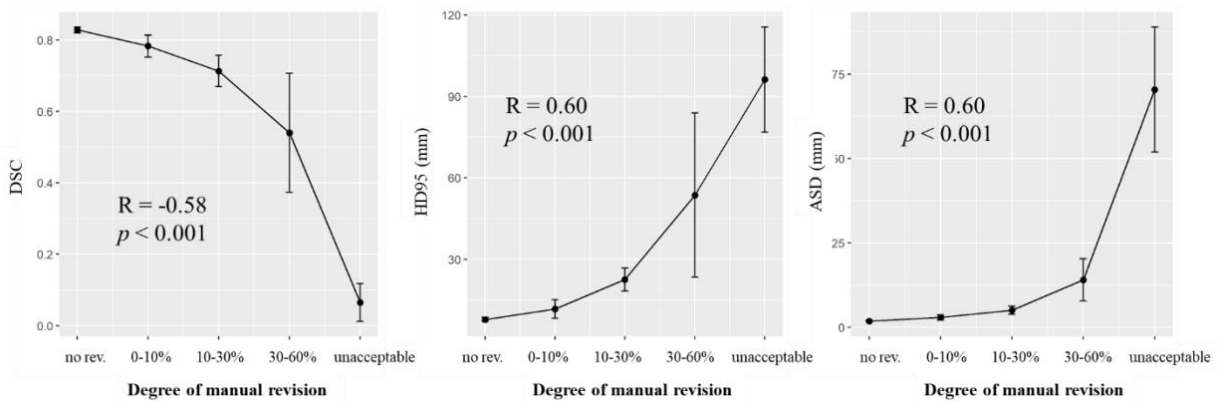

Fig. 4 Expert assessment of manual revision degree of the deep model predicted contours. Table in the top row summarized the mean and 95% confidence interval (CI) of Dice similarity coefficient (DSC), 95% Hausdorff distance (HD95) and average surface distance (ASD) stratified by different degrees of manual revision. The correlations between the mean of DSC, HD95, ASD and the degree of manual revision were plotted in the bottom row. Spearman correlation coefficient showed that DSC and degree of manual revision were correlated (R=-0.58, $p<0.001$). Same correlation was observed for the HD95 and ASD (HD95: R=0.60, $p<0.001$, ASD: R=0.60, $p<0.001$). Degree of manual revision was defined as the percentage of GTV slices that needed modification for clinical acceptance. pCT = treatment planning CT.

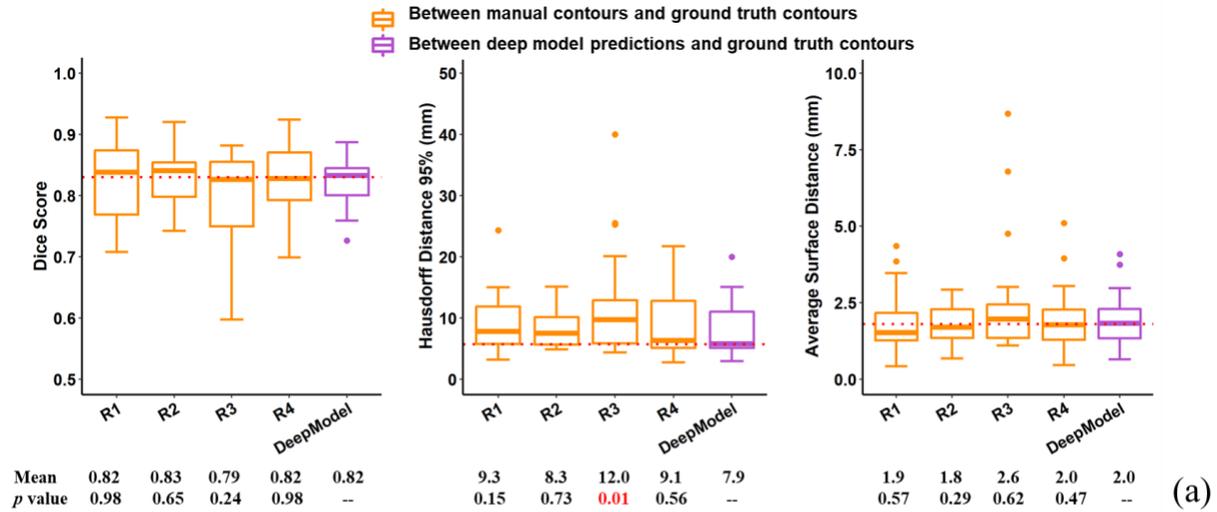

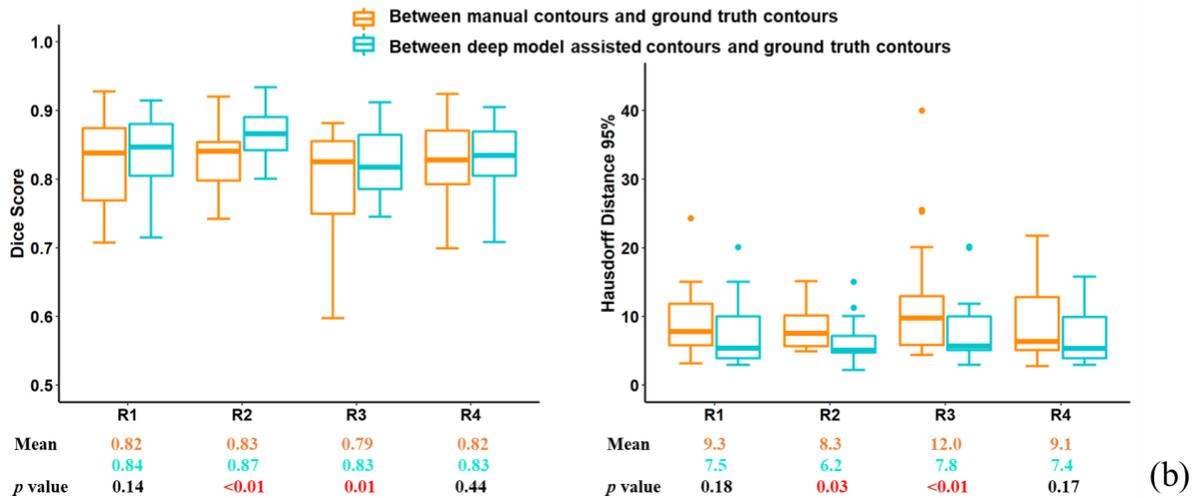

Fig. 5. Results of multi-user evaluation. (a): boxplot of Dice similarity coefficient (DSC), 95% Hausdorff distance (HD95) and average surface distance for the comparison of manual contours of four radiation oncologists with our pCT-based deep model predicted contours. Dotted lines indicate the median DSC, HD95 and ASD of our pCT model performance. (b): Comparison of DSC and HD95 between 2nd-time deep learning assisted contours with those of 1st-time manual contours. R1 to R4 represent the 4 radiation oncologists involved in the multiuser evaluation. DeepModel is our pCT model.

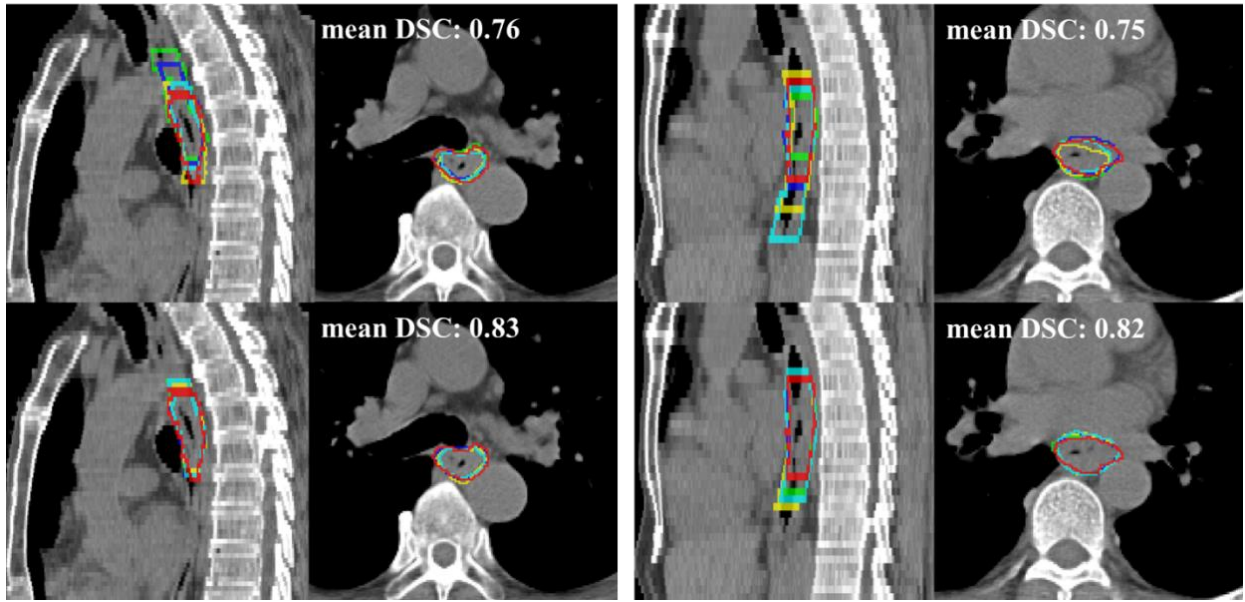

Fig. 6. Two qualitative examples (left and right) in sagittal and axial views of comparison between the 1st-time manual contour (top row) and 2nd-time deep learning assisted contours (bottom row). Red color is the ground truth contour, while green, blue, yellow and cyan represent the other four radiation oncologists' contours. The average Dice similarity coefficient (DSC) of 4 radiation oncologists for their 1st-time manual contour is 0.76 and 0.75 to the two examples, respectively. The DSC performance improved to 0.83 and 0.82 for their 2nd-time contour with assistance from the deep learning predictions.

# Supplemental Materials

*Imaging protocols:*

Treatment planning CT (pCT) scans from the 4 institutions were acquired with scanners from various vendors, including Siemens and GE healthcare. Slice thickness and in-plane pixel size ranged from 0.78×0.78×1.25mm to 1.37×1.37×5mm (median resolution 0.98×0.98×5mm). The FDG-PET/CT image pairs of 252 patients in institution 1 were acquired using GE Discovery ST scanner (Milwaukee, WI), or Siemens Biograph mCT PET/CT scanner (Hoffman Estates, IL). Before PET acquisition a diagnostic quality CT was performed with 5 mm slice width and average 1mm in-plane pixel size. The PET images were reconstructed with CT-based attenuation correction and the axial spatial resolutions of PET were 4.80 and 2.16 mm for the Discovery ST and Biograph mCT scanners, respectively.

*PET/CT to pCT registration:*

Direct PET to pCT registration can lead to large errors because of the completely different modalities (1). To overcome this challenge, we register the diagnostic CT (accompanying the PET) to pCT and use the resulting deformation field to align the PET to pCT. This intra-patient CT-based registration is a much feasible task, where many deformable registration algorithms have demonstrated excellent results. In this work, we used the Dense Displacement Sampling algorithm (DEEDS) (2), which achieved the leading performance of an average target registration error $\leqslant$ 2mm in a pulmonary registration challenge (3).

However, to achieve robust performance, diagnostic CT and pCT require a reasonable rigid initialization to manage their pose, scanning range, and respiratory differences (Fig. 3 (a)) before applying the deformable algorithm. To accomplish this step of registration, we used 3D mass centers of lungs in two CT scans as the initial matching position, where 3D lung masks were automatically segmented using the method of (4). This led to a reliable initial alignment for the chest and upper abdominal regions, ensuring the success of deformable registration process. The resulting deformation field is subsequently applied to PET scan and aligned it to the respective pCT scan. A registration example is shown in Fig. A1.

*Two-streamed GTV segmentation workflow:*

A two-streamed 3D deep learning model for esophageal GTV segmentation was developed, which had the flexibility to segment the GTV using only pCT, or pCT+PET/CT when PET/CT is available. One pCT-based deep network was trained using the only pCT to segment the GTV (denoted as pCT-stream). The other pCT+PET stream consisted of an early fusion network followed by a late fusion network to segment the GTV, which utilizes the complementary information in pCT and PET scans. More details of this method can be found in (5).

*Progressive semantically-nested network (PSNN) architecture:*

The detailed network architecture of PSNN is shown in Fig. A2. It incorporates the strengths of both UNet (6) and PHNN (7) by using deep supervision (8) to progressively propagate high-level semantic features to lower-levels, but higher resolution features. We will deposit all computer codes used for modeling and inference in GitHub once the work is accepted.

*Training and inference details:*

We used a 3D volume of interest (VOI) patch-based fashion to train all the deep networks in the two-streamed method. To generate the 3D training samples, we extracted 96×96×64 sub-volumes in two manners: (1) To ensure enough VOIs with positive GTV content, we randomly extract VOIs centered within the ground truth GTV masks. (2) To obtain sufficient negative examples, we randomly sample 20 VOIs from the rest of the whole volume. We further applied extensive data augmentation, e.g., horizontal flipping, random rotations in the x-y plane within ±10 degrees, intensity scaling with a ratio between [0.75, 1.25], Gaussian noise with zero mean and (0, 0.1) variance. The optimizer is stochastic gradient descent with the Polynomial learning rate policy. The initial learning rate is 0.01 and a Nesterov momentum of 0.99. Batch size is set to 12 for all networks. The pCT and early fusion deep network were trained for 150 epochs to convergence, while the late fusion deep network was trained for 50 epochs. For inference, we first cropped out the lung xy ranges using the 3D lung mask, then, 3D sliding windows with sub-volumes of 96×96×64 and strides of 64×64×32 voxels were used. The probability maps of sub-volumes were aggregated to obtain the whole volume prediction. We implemented our model using Pytorch and Titan-V GPU.

*Supplemental References*

*Supplemental Tables*

Table A1. Subject and imaging characteristics in the multi-user study.

| Characteristics | Multiuser study Institutions 2-4 (n = 20) |
|---|---|
| Sex | … |
|    Male | 18 (90%) |
|    Female | 2 (10%) |
| Diagnostic age | 69 [61-79] |
| Clinical T stage | … |
|    cT2 | 4 (20%) |
|    cT3 | 10 (50%) |
|    cT4 | 6 (30%) |
| Tumor location | … |
|    Cervical | 2 (10%) |
|    Upper third | 3 (34%) |
|    Middle third | 12 (60%) |
|    Lower third | 11 (55%) |
| BMI | … |
|    < 18.5 | 7 (35%) |
|    18.5 – 23.9 | 11 (55%) |
|    > 24 | 2 (10%) |
| Imaging available | … |
|    pCT | 20 (100%) |
|    PET/CT | 0 (0%) |

Note: patients may have tumors located across multiple esophagus regions, hence, total numbers summed at various tumor locations for the entire and sub-institution cohorts are greater than the corresponding total patient numbers. Age is presented as median and [interquartile range]. pCT = treatment planning CT.

Table A2. Contour accuracy comparison to pCT-based DenseUNet method (9) in external multi-institutional testing.

| Metrics | Our pCT deep model | DenseUNet model | p value |
|---|---|---|---|
| DSC: mean (95% CI) | 0.80 (0.78, 0.81) | 0.75 (0.73, 0.77) | <0.001 |
| HD95 (mm): mean (95% CI) | 11.8 (10.1, 13.4) | 15.5 (13.0, 18.1) | <0.001 |
| ASD (mm): mean (95% CI) | 2.8 (2.4, 3.2) | 4.3 (3.5, 5.2) | <0.001 |

Note: Wilcoxon matched-pairs signed rank test was used to compare the performance of our developed model and the DenseUNet model (9). pCT = treatment planning CT. DSC = Dice similarity coefficient, HD95 = 95% Hausdorff distance, ASD = average surface distance. CI = confidence internal.

Table A3. Experts' assessment of degrees of manual revision in the subgroup analysis stratified by clinical T stages and tumor locations.

| | Institution 2-4 (external multi-institutional testing) | | | | |
|---|---|---|---|---|---|
| Input images | pCT | | | | |
| Degree of manual revision | no revision (n=220) | >0-10% (n=49) | >10%-30% (n=42) | >30%-60% (n=10) | unacceptable (n=33) |
| Clinical T stage | | | | | |
|   cT2 (n=74) | 38 (52%) | 4 (5%) | 6 (8%) | 3 (4%) | 23 (31%) |
|   cT3 (n=177) | 125 (70%) | 26 (15%) | 17 (10%) | 4 (2%) | 5 (3%) |
|   cT4 (n=103) | 57 (55%) | 19 (18%) | 19 (18%) | 3 (3%) | 5 (5%) |
| Tumor location | | | | | |
|   Cervical (n=60) | 37 (62%) | 7 (12%) | 10 (17%) | 2 (3%) | 4 (6%) |
|   Upper third (n=143) | 82 (57%) | 21 (15%) | 25 (17%) | 4 (3%) | 11 (8%) |
|   Middle third (n=178) | 107 (60%) | 31 (18%) | 18 (10%) | 6 (3%) | 16 (9%) |
|   Lower third (n=70) | 45 (64%) | 5 (7%) | 10 (14%) | 4 (6%) | 6 (9%) |

Note: the $\chi^2$ test was used to compare the difference in degrees of manual revision between subgroups. A significantly higher percentage of patients required minor (0-30% slice revision) or no revision in advanced cT3 and cT4 stages as compared to that in early cT2 stage (95% and 91% vs 65%, $p<0.01$). cT2 = clinical T-stage 2, cT3 = clinical T-stage 3, and cT4 = clinical T-stage 4. pCT = treatment planning CT.

Table A4. Contouring accuracy comparison of four radiation oncologists for their 1st time manual delineation.

| Evaluation metrics | R1 | R2 | R3 | R4 | Deep model |
|---|---|---|---|---|---|
| DSC | … | … | … | … | … |
|   mean (95% CI) | 0.82 (0.79, 0.85) | 0.83 (0.80, 0.85) | 0.79 (0.76, 0.83) | 0.82 (0.79, 0.85) | 0.82 (0.80, 0.84) |
|   *p* value | 0.98 | 0.65 | 0.24 | 0.98 | -- |
| HD95 (mm) | … | … | … | … | … |
|   mean (95% CI) | 9.3 (6.9, 11.7) | 8.3 (6.8, 9.8) | 12.0 (7.7, 16.3) | 9.1 (6.5, 11.7) | 7.9 (5.7, 10.0) |
|   *p* value | 0.15 | 0.73 | **0.01** | 0.56 | -- |
| ASD (mm) | … | … | … | … | … |
|   mean (95% CI) | 1.9 (1.4, 2.4) | 1.8 (1.5, 2.1) | 2.6 (1.6, 3.5) | 2.0 (1.4, 2.5) | 2.0 (1.6, 2.4) |
|   *p* value | 0.57 | 0.29 | 0.62 | 0.47 | -- |

Note: R1 to R4 represent the 4 radiation oncologists involved in the multiuser evaluation. The Wilcoxon matched-pairs signed rank test was used to compare the performance of our developed model and the four radiation oncologists. CI = confidence internal, DSC = Dice similarity coefficient, HD95 = 95% Hausdorff distance, ASD = average surface distance.

Table A5. Contouring accuracy comparison of four radiation oncologists between their 1st time manual delineation and 2nd time deep learning assistant delineation in terms of DSC and HD95.

| Radiation oncologists | DSC mean (95% CI) | p value | HD95 (mm) mean (95% CI) | p value |
|---|---|---|---|---|
| R1 |  | 0.14 |  | 0.18 |
| 1st time manual contour | 0.82 (0.79, 0.85) |  | 9.3 (6.9, 11.7) |  |
| 2nd time assisted contour | 0.84 (0.82, 0.86) |  | 7.5 (5.3, 9.6) |  |
| R2 |  | **<0.01** |  | **0.03** |
| 1st time manual contour | 0.83 (0.80, 0.85) |  | 8.3 (6.8, 9.8) |  |
| 2nd time assisted contour | 0.87 (0.85, 0.88) |  | 6.2 (4.6, 7.7) |  |
| R3 |  | **0.01** |  | **<0.01** |
| 1st time manual contour | 0.79 (0.76, 0.83) |  | 12.0 (7.7, 16.3) |  |
| 2nd time assisted contour | 0.83 (0.80, 0.85) |  | 7.8 (5.4, 10.1) |  |
| R4 |  | 0.44 |  | 0.17 |
| 1st time manual contour | 0.82 (0.79, 0.85) |  | 9.1 (6.5, 11.7) |  |
| 1st time manual contour | 0.83 (0.81, 0.86) |  | 7.4 (5.3, 9.5) |  |
| Average of R1 to R4 |  | **<0.001** |  | **<0.001** |
| 1st time manual contour | 0.82 (0.80, 0.83) |  | 9.7 (8.3, 11.1) |  |
| 1st time manual contour | 0.84 (0.83, 0.85) |  | 7.2 (6.2, 8.2) |  |

Note: R1 to R4 represent the 4 radiation oncologists involved in the multiuser evaluation. The Wilcoxon matched-pairs signed rank test was used to compare the radiation oncologist's performance of the 1st time manual contouring and 2nd time assisted contouring. DSC = Dice similarity coefficient, HD95 = 95% Hausdorff distance.

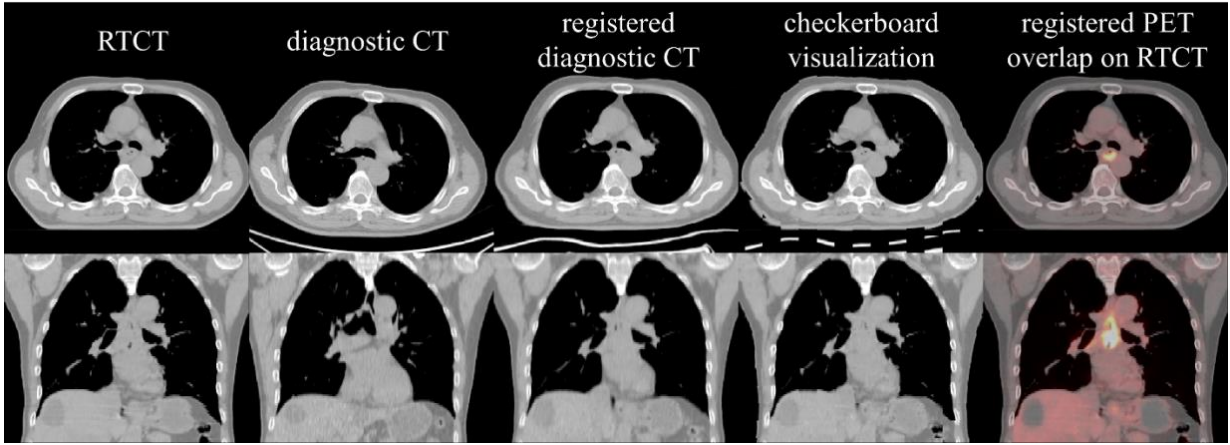

Fig. A1. An example of the deformable registration results for a patient in axial and coronal views. From left to right are the pCT image; the PET-CT image (accompanying PET) before and after the registration, respectively; a checkerboard visualization of the pCT and registered diagnostic CT images; and finally, the overlapped PET image, transformed using the diagnostic CT deformation field, on top of the pCT. pCT = treatment planning CT.

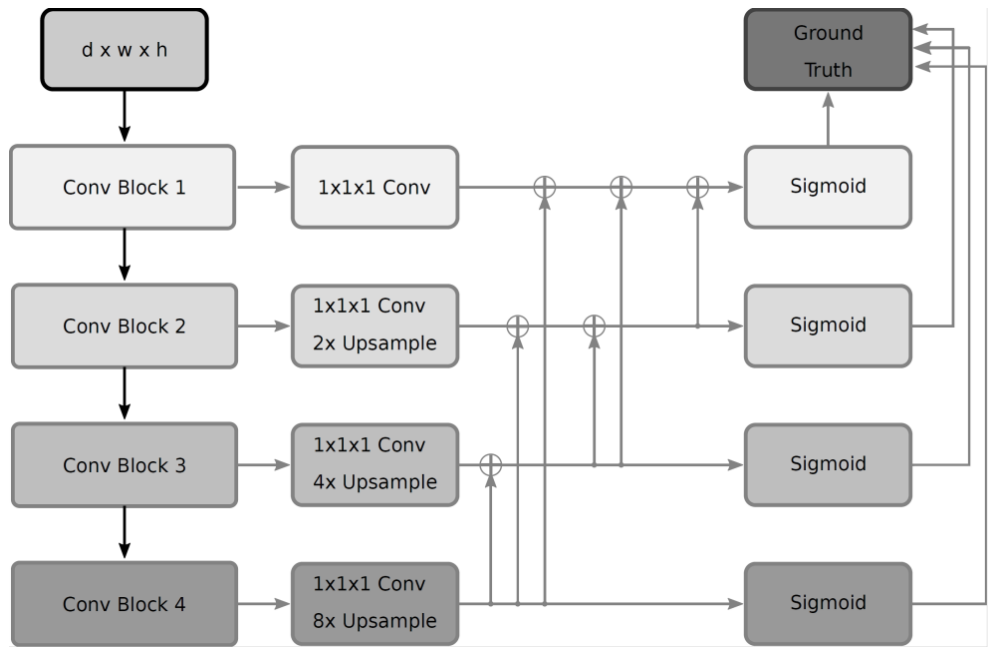

Fig. A2. Illustration of the 3D Progressive semantically-nested network (PSNN) model, which employs deep supervision at different scales within a parameter-less high-to-low level image segmentation decoder. Four 3D convolutional blocks are applied. The first two and last two blocks are composed of two and three 3×3×3 convolutional+BN+ReLU layers, respectively.

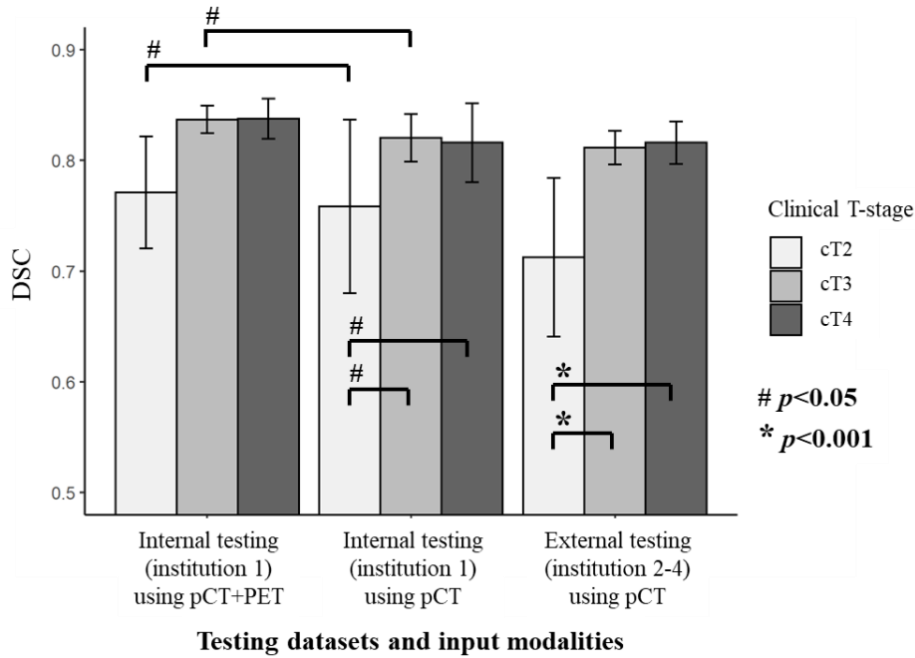

Fig. A3. Subgroup analysis of the deep model performance stratified by the clinical T-stage in both internal and external testing datasets. Manning-Whitney U test was used to compare the DSC of different clinical T-stage patients in each testing dataset and input modality. The Wilcoxon matched-pairs signed rank test was used to compare the DSC between the model using pCT and that using the pCT+PET in internal testing. Using pCT as input, DSC of cT2 patients yielded significantly lower values as compared to that of cT3 and cT4 patients in both internal (mean DSC: 0.76 vs 0.82, 0.82, $p<0.05$) and external (mean DSC: 0.71 vs 0.81, 0.82, $p<0.001$) evaluation. This phenomenon was less prominent after adding PET as additional input. In internal evaluation, using pCT+PET as input, DSC of patients of cT2 and cT3 had significantly improved performance ($p<0.05$) over those using pCT alone as input. cT2 = clinical T-stage 2, cT3 = clinical T-stage 3, and cT4 = clinical T-stage 4. pCT = treatment planning CT, DSC = Dice coefficient similarity.

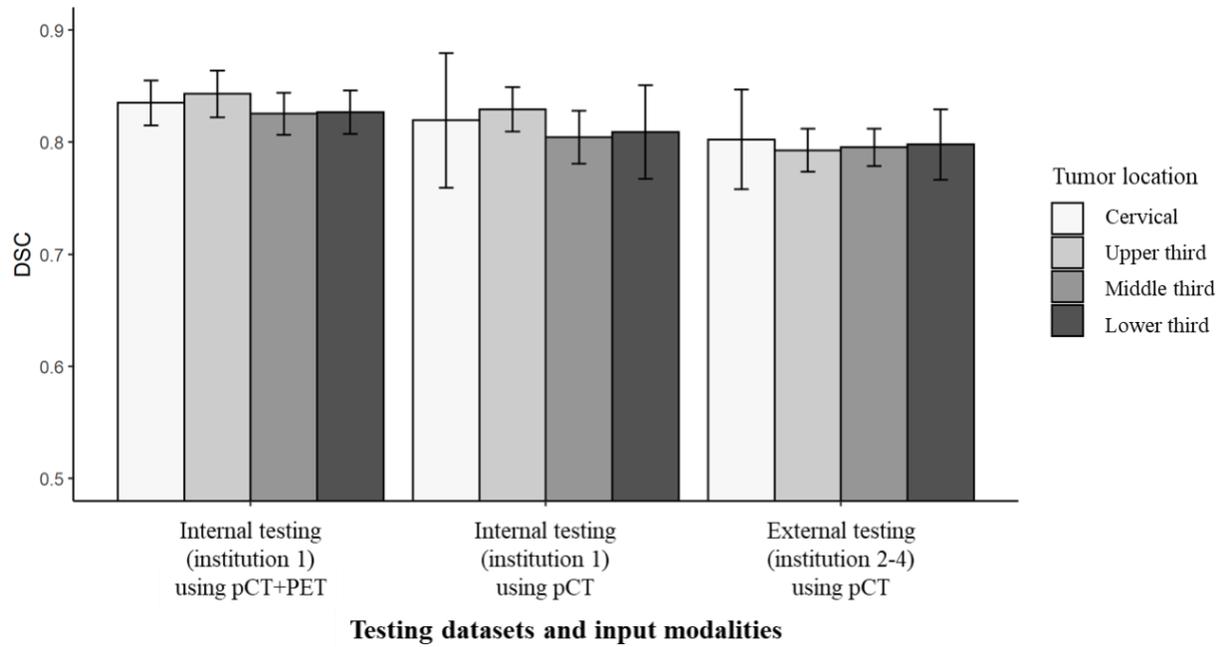

Fig. A4. Subgroup analysis of the deep model performance stratified by the tumor location in both internal and external testing datasets. pCT = planning CT, DSC = Dice coefficient similarity. Multiple linear regression with stepwise model selection was used to compare the DSC at different tumor locations, since a large tumor may locate across multiple esophagus regions. No significant differences were observed for the deep model performance at four tumor locations in both internal and external testing datasets or using pCT and pCT+PET as input images. This demonstrated the robustness of our deep model at different tumor locations. pCT = treatment planning CT, DSC = Dice coefficient similarity.